\documentclass[runningheads]{llncs}
\usepackage[T1]{fontenc}
\usepackage{graphicx}
\usepackage{booktabs}
\usepackage[misc]{ifsym}

\usepackage{hyperref}
\usepackage{float}
\usepackage{csquotes}
\usepackage[utf8]{inputenc}
\usepackage[T1]{fontenc}

\begin{document}

\title{T-REX: Table -- Refute or Entail eXplainer}

\titlerunning{T-REX: Table -- Refute or Entail eXplainer}

\author{Tim Luka Horstmann (\Letter) \and Baptiste Geisenberger \and Mehwish Alam \orcidID{0000-0002-7867-6612}}

\authorrunning{T.L. Horstmann et al.}
\toctitle{T-REX: Table – Refute or Entail eXplainer}
\tocauthor{Tim Luka Horstmann, Baptiste Geisenberger, Mehwish Alam}

\institute{Institut Polytechnique de Paris, France \\ \email{\{tim.horstmann, geisenberger\}@ip-paris.fr, mehwish.alam@telecom-paris.fr}
}

\maketitle              

\begin{abstract}
Verifying textual claims against structured tabular data is a critical yet challenging task in Natural Language Processing with broad real-world impact. While recent advances in Large Language Models (LLMs) have enabled significant progress in table fact-checking, current solutions remain inaccessible to non-experts. We introduce T-REX (Table – Refute or Entail eXplainer), the first live, interactive tool for claim verification over multi\-modal, multilingual tables using state-of-the-art instruction-tuned reasoning LLMs. Designed for accuracy and transparency, T-REX empowers non-experts by providing access to advanced fact-checking technology. The system is openly available online.
\newline
Online Demo: \url{https://t-rex.r2.enst.fr}
\newline
Demo (video): \url{https://www.youtube.com/watch?v=HHIxVCOT8X0}
\newline
Github: \url{https://github.com/TimLukaHorstmann/T-REX}


\end{abstract}

\section{Introduction}
Table fact-checking is the Natural Language Processing task of classifying textual claims as \textit{entailed} or \textit{refuted} based on information contained in a table. Unlike traditional fact-checking, which primarily deals with unstructured text, this task demands a combination of capabilities, including linguistic understanding, symbolic reasoning, and numerical computation. For example, a BBC headline once claimed \textit{\enquote{Germany: Migrants \enquote{may have fueled violent crime rise}}}, despite official crime statistics showing no such trend.\footnote{\href{https://www.bbc.com/news/world-europe-42557828}{BBC News article}; \href{https://www.bka.de/EN/CurrentInformation/Statistics/PoliceCrimeStatistics/2020/pcs2020.html}{Federal Criminal Police Office – Police Crime Statistics 2020}} With the right tools, such discrepancies could be automatically detected by reasoning over the corresponding tabular data. Since tables are a primary format for structured data across domains like journalism, finance, policymaking, and science, robust and accessible table fact-checking systems are increasingly vital to combat misinformation. 

Advances in Large Language Models (LLMs) have driven significant progress in table fact-checking. RePanda~\cite{chegini_repanda_2025} translates claims into executable queries for explainable verification, while DATER~\cite{ye_large_2023} and ARTEMIS-DA~\cite{hussain_artemis-da_2024} decompose tables and claims into structured reasoning steps, surpassing human performance on the Tab\-Fact data\-set~\cite{chen_tabfact_2020}. Despite these advances, most LLM-based verification systems remain confined to offline evaluation, lack interactivity, and require technical expertise --- ultimately restricting accessibility for non-expert users.

Some tools partially address these limitations. OpenTFV~\cite{gu_opentfv_2022} retrieves tables from corpora and verifies claims with table-aware models and textual explanations, while Aletheia~\cite{fu_data_2024} maps claims to structured datasets for article fact-checking with visualized evidence presentation. However, both assume predefined datasets, rely on earlier-generation encoder-only or proprietary LLMs, and are not designed to operate with arbitrary, user-provided tables as primary inputs.

To bridge these gaps, we introduce T-REX (Table – Refute or Entail eXplainer), the first tool for real-time table fact-checking using state-of-the-art (SOTA) instruction-tuned reasoning LLMs. T-REX prioritizes accuracy through a robust verification pipeline, accessibility via a multilingual, multimodal, and user-friendly interface, and interpretability through transparent reasoning outputs. T-REX extends table fact-checking beyond expert interactions into real-world applications, where accurate and accessible claim verification is critical.

\section{System Architecture and Design} \label{sec:t_rex}

\begin{figure}[H]
\includegraphics[width=\textwidth]{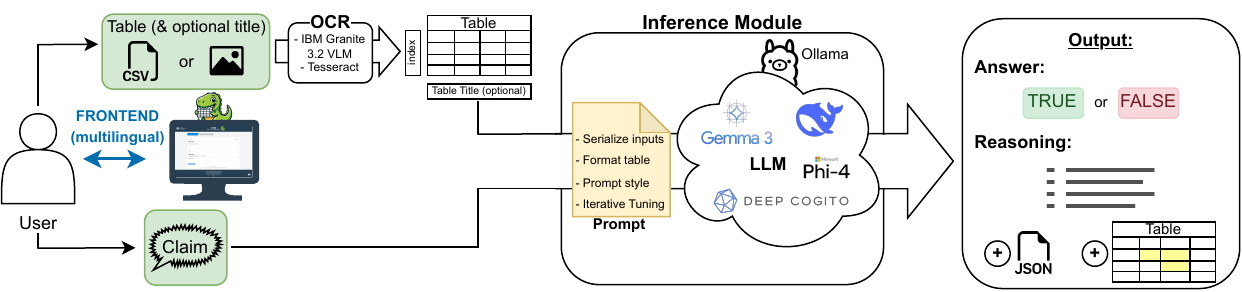}
\caption{T-REX system overview. Users provide tables and claims via the frontend, while the backend handles OCR, preprocessing, and LLM-based verification.}
\label{fig:system_overview}
\end{figure}

T-REX integrates a modular framework for table processing, LLM inference, Optical Character Recognition (OCR), and a web-based frontend for interactive verification, visualization, and real-time explanations (see Figure~\ref{fig:system_overview}).

\subsection{Table Processing and Verification Pipeline}
\label{ssec:main_pipeline}

T-REX supports table input via CSV files or OCR-extracted images, using IBM’s Gra\-nite 3.2 vision model for high accuracy or Tesseract as a faster alternative. Tables are preprocessed for robustness (unifying delimiters, cleaning contents, and standardizing structure) and interpretability (by injecting a synthetic \texttt{row\_index} column to aid the model's cell referencing). A claim, and optionally a table title for additional context, are then paired with the table and passed to the inference engine for verification.

To guide the design of our inference pipeline, we conducted extensive offline experiments on the TabFact test set~\cite{chen_tabfact_2020}. We compared three strategies: (1) direct prompting for claim verification, (2) prompting the model to generate code (Python or SQL) that returns \textit{True} or \textit{False}, and (3) Retrieval-Augmented Generation, embedding the claim and the table (row-, column-, or cell-wise) and selecting relevant table content based on cosine similarity. Each approach was evaluated across five open-source LLMs (Llama 3.2 (3B), Mis\-tral (7B), DeepSeek-R1 (8B and 32B), and Phi-4 (14B)), four formats for representing the table within the prompt (Markdown, HTML, JSON, and naturalized text), and two prompting techniques (zero-shot and chain-of-thought (CoT)). We iteratively refined the prompt design through manual evaluation, drawing on best practices from prior work. We found that concise, directive, and clear prompts consistently led to more reliable outputs. Detailed results for (1) are available in the online demo; final results for (1-3) on our GitHub.

Our best-performing setup, Phi-4 with naturalized table formatting and CoT prompting, achieved 89\% accuracy, surpassing prior methods like Re\-Pan\-da on TabFact~\cite{chegini_repanda_2025,chen_tabfact_2020}. Beyond accuracy, direct prompting with CoT enhances interpretability (Section~\ref{ssec:frontend}) by allowing us to guide the model to output its reasoning, followed by a JSON object containing the final verdict and self-identified relevant table cells. We apply post-processing to the model's output to robustly extract the JSON, using fallback mechanisms to recover the verdict in case the JSON parsing fails. T-REX currently supports four open-source SOTA LLMs, balancing structured reasoning abilities with GPU constraints: Phi-4 (14B), Co\-gi\-to v1 Preview (8B), DeepSeek-R1-Distill-Qwen-7B (7B), and Gemma 3 (4B).

\subsection{Interface Design and User Features}
\label{ssec:frontend}

\begin{figure}[t]
\centering
\includegraphics[width=\textwidth]{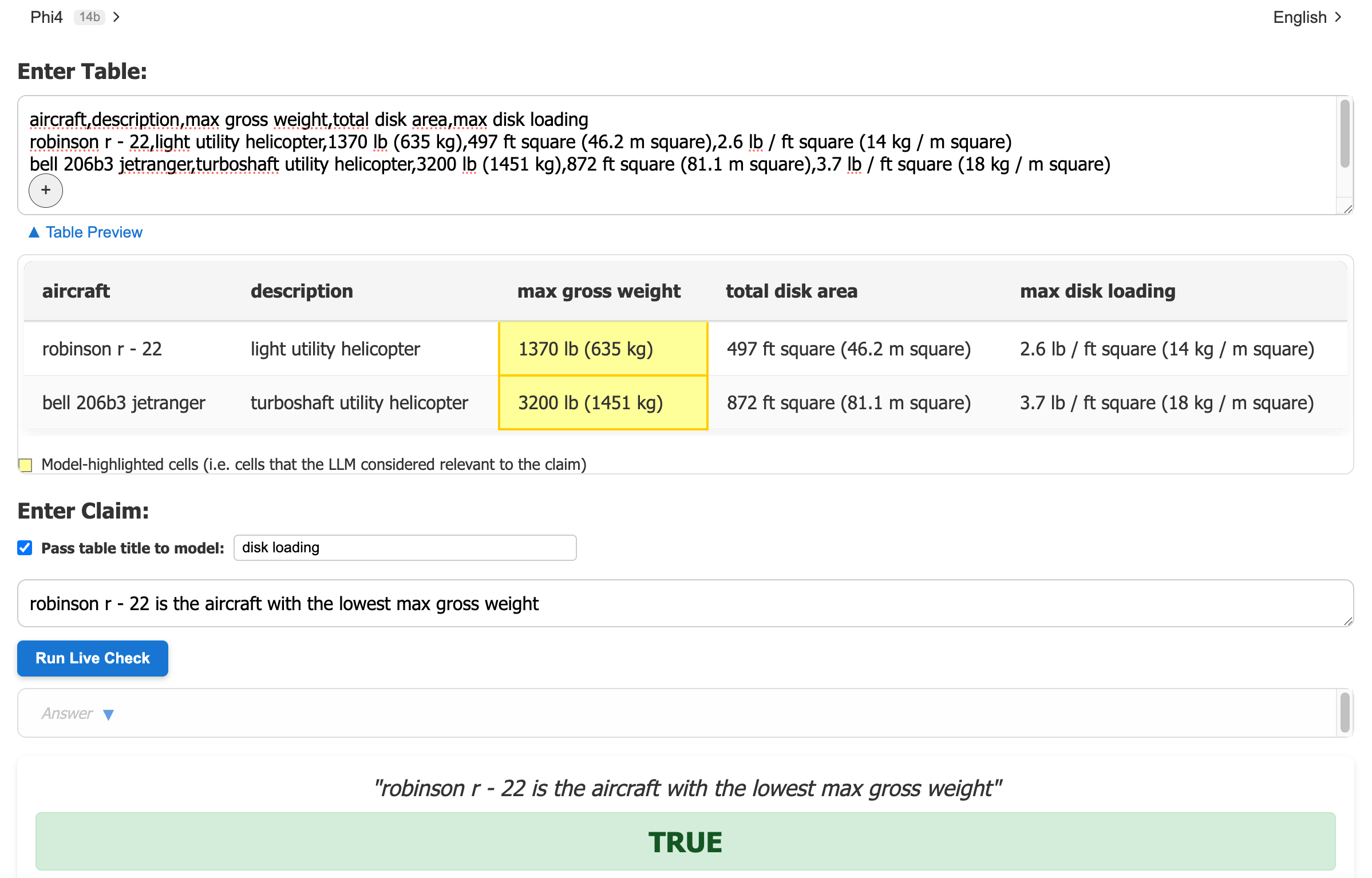}
\caption{T-REX interface after claim verification. Relevant cells are highlighted for interpretability and reflect the model's self-identified reasoning steps. The model’s reasoning is accessible via dropdown menus.}
\label{fig:UI}
\end{figure}

The T-REX interface (Figure~\ref{fig:UI}) prioritizes accessibility and transparency. Users initiate the verification process by uploading or pasting a table (CSV or image) and a claim (Section~\ref{ssec:main_pipeline}), or by selecting a TabFact~\cite{chen_tabfact_2020} example, choosing an LLM, and clicking the ``Run Live Check'' button.  For the hybrid reasoning model Cogito, users can enable the ``Deep Thinking'' mode to allow deeper reasoning. T-REX streams its reasoning in real time, generates a verdict, and highlights relevant table cells within the table preview for improved interpretability.

To further enhance usability, T-REX offers multilingual input and output in eight languages, exportable JSON outputs, Wikipedia previews for the original source pages of TabFact examples~\cite{chen_tabfact_2020}, intuitive navigation via dropdown selectors, and dark mode. T-REX also automatically renders all table inputs as editable CSV within the interface, synchronized with a live table preview for enhanced visualization. This feature is particularly useful for OCR inputs, where recognition errors can be quickly identified and manually corrected by the user.

\subsection{Implementation Details}
\label{ssec:implementation_details}
T-REX is built on a modular FastAPI backend that orchestrates LLM inference, OCR processing, and data management via asynchronous HTTP calls. Ollama serves both LLM inference and OCR tasks, while SlowAPI enforces request rate limiting to maintain robustness under load. T-REX currently runs on two NVIDIA GTX 1080 GPUs, and its modular design makes it easy to scale to larger LLMs as hardware improves. All inference and data handling is performed in memory, ensuring that no user data is stored or transmitted externally.

\section{Conclusion}
\label{sec:conclusion}
This paper introduced T-REX, the first live, interactive, and domain-agnostic table fact-checking tool powered by SOTA open-source instruction-tuned reasoning LLMs. Designed to facilitate access to advanced claim verification for non-experts, T-REX provides accurate and interpretable verification in real time, without storing or exposing any user data externally. In the future, we plan to extend T-REX to integrate precise numerical computation, structured claim decomposition, and support for additional table-centric tasks such as table completion, semantic table retrieval, and cell-level evidence justification. 

\begin{credits}
\subsubsection{\discintname}
The authors have no competing interests to declare that are relevant to the content of this article.
\end{credits}


%
%
%
%

\end{document}